# Real-time Webcam Heart-Rate and Variability Estimation with Clean Ground Truth for Evaluation ♡

Amogh Gudi [V,D,*,★] , Marian Bittner [V,D,*,★] , and Jan van Gemert [D]

V  Vicarious Perception Technologies (VicarVision), Amsterdam, The Netherlands; www.vicarvision.nl
D  Delft University of Technology, Delft, The Netherlands; www.tudelft.nl
*  Correspondence: {amogh, marian}@vicarvision.nl (A.G., M.B.); j.c.vangemert@tudelft.nl (J.v.G.)
♡  This paper is an extended version of our paper published in IEEE ICCVW 2019, Seoul, Korea.
★  These authors contributed equally to this work.

**Featured Application:** Contactless pervasive health monitoring including cognitive load/stress analysis.

**Abstract:** Remote photo-plethysmography (rPPG) uses a camera to estimate a person's heart rate (HR). Similar to how heart rate can provide useful information about a person's vital signs, insights about the underlying physio/psychological conditions can be obtained from heart rate variability (HRV). HRV is a measure of the fine fluctuations in the intervals between heart beats. However, this measure requires temporally locating heart beats with a high degree of precision. We introduce a refined and efficient real-time rPPG pipeline with novel filtering and motion suppression that not only estimates heart rates, but also extracts the pulse waveform to time heart beats and measure heart rate variability. This unsupervised method requires no rPPG specific training and is able to operate in real-time. We also introduce a new multi-modal video dataset, VicarPPG 2, specifically designed to evaluate rPPG algorithms on HR and HRV estimation. We validate and study our method under various conditions on a comprehensive range of public and self-recorded datasets, showing state-of-the-art results and providing useful insights into some unique aspects. Lastly, we make available CleanerPPG, a collection of human-verified ground truth peak/heart-beat annotations for existing rPPG datasets. These verified annotations should make future evaluations and benchmarking of rPPG algorithms more accurate, standardized and fair.

**Keywords:** remote photoplethysmography; heart rate variability; real-time; unsupervised; clean ground truth

## 1. Introduction

Human vital signs like heart rate, blood oxygen saturation and related physiological measures can be measured using a technique called photo-plethysmography (PPG). This technique involves optically monitoring light absorption in tissues that are associated with blood volume changes. Typically, this is done via a contact sensor attached to the skin surface [1]. Such contact sensors can detect the underlying vital signs quite reliably owing to their proximity to the subject, and therefore have applications in critical areas like patient monitoring. However, the ability to obtain such measurements remotely via a camera/webcam, albeit less accurately, can enable applications outside the medical domain (e.g. affective computing, human-computer interaction), where contact sensors are not feasible. Remote Photo-plethysmography (rPPG) detects the blood volume pulse remotely by tracking changes in the skin reflectance as observed by a camera [2,3]. In this paper we present a novel framework for extracting heart rate (HR) and heart rate variability (HRV) from the face. This work is an extension of the work done in [4].

Vital signs from videos. The process of rPPG essentially involves two steps: detecting and tracking the skin colour changes of the subject, and analysing this signal to compute measures like heart rate, heart rate variability and respiration rate. Recent advances in computer video, signal processing, and machine learning have improved the performances of rPPG techniques significantly [2]. Current state-of-the-art methods are able to leverage image processing by supervised deep neural networks to robustly select skin pixels within an image and perform HR estimation [5,6]. However, this reliance upon heavy machine learning (ML) processes has two primary drawbacks: (i) it necessitates rPPG specific fully supervised training of the ML model, thereby requiring collection of large training sets; (ii) complex models can require significant computation time on CPUs and thus can potentially add a bottleneck in the pipeline and limit real-time utility. Since rPPG analysis is based on a signal processing task, the use of an end-to-end trainable system with no domain knowledge leaves room for improvement in efficiency (e.g., we *know* that pulse signal is embedded in average skin colour changes [3,7,8], but the machine learning system has to *learn* this). We introduce an efficient unsupervised rPPG pipeline that performs the full rPPG analysis in real-time. This method achieves state-of-the-art results without needing any rPPG related training. This is achieved via extracting regions of interest robustly by 3D face modelling, and explicitly tracking and reducing the influence of head movement to filter the signal.

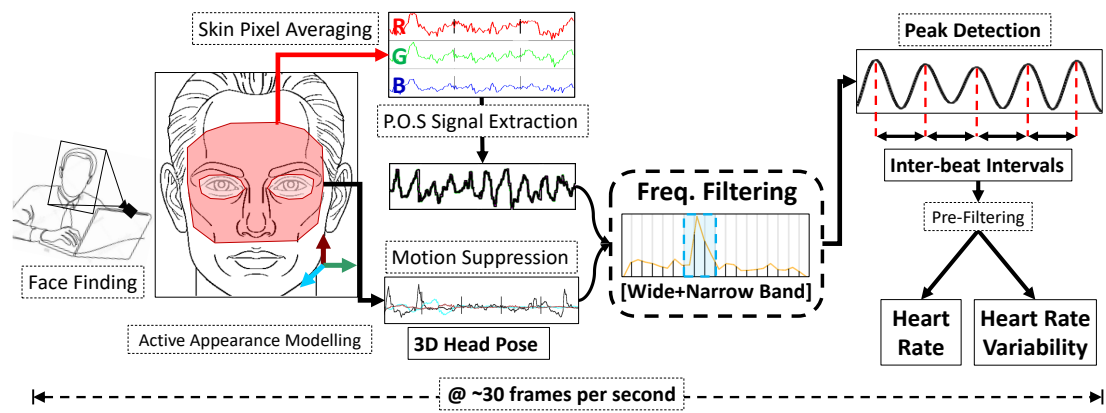

**Figure 1.** An overview of the proposed heart rate and heart rate variability estimation pipeline (left to right). The face in captured webcam images are detected and modelled to track the skin pixels in region of interest. A single 1-D signal is extracted from the spatially averaged values of these pixels over time. In parallel, 3-D head movements are tracked and used to suppress motion noise. An FFT based wide and narrow band filtering process produces a clean pulse waveform from which peaks are detected. The inter-beat intervals obtained from these peaks are then filtered and used to compute heart rate and heart rate variability. The full analysis can be performed in real time on a CPU.

Heart rate variability. While heart rate is a useful output from a PPG/rPPG analysis, finer analysis of the obtained blood volume pulse (BVP) signal can yield further useful measures. One such measure is heart rate variability (HRV): an estimate of the variations in the time-intervals between individual heart beats. This measure has utility in providing insights into the physiological and psychological state of a person (stress levels, anxiety, etc.). While traditionally this measure is obtained based on observation over hours [9], short and ultra-short duration ($\leq 5$ mins) HRV are also being studied [10]. Our experiments focus on obtaining ultra-short HRV measure as a proof-of-concept/technology demonstrator for longer duration applications.

The computation of heart rate variability depends on the time-variations between heart beats, and it therefore requires temporally locating the heart beats with a high degree of accuracy. Unlike HR estimation, where errors in opposite directions average out, HRV analysis is sensitive to even small artefacts and all errors add up to strongly distort the final measurement. Thus, estimating HRV is a challenging task for rPPG and this has received relatively little focus in literature. Our method extracts a clean BVP signal from the input via a two-step wide and narrow band frequency filter to accurately time heart beats and estimate heart rate variability. An overview of our method is illustrated in Figure 1.

State of datasets. As the field of remote photo-plethysmography receives advances and the accuracies of rPPG methods improve, the demand for thorough, challenging and realistic datasets arise. Datasets originally created for alternate uses (e.g. psychological studies) often get re-purposed for rPPG analysis, which has unintended drawback (video compression, occlusion, etc.) [11,12]. A large proportion of research in this field end up using self-recorded private datasets [13], due to which the results cannot be directly compared with prior work. These factors hinder proper development, evaluation and benchmarking of rPPG methods. To help alleviate this, we introduce a new publicly available multi-modal video dataset specifically designed to aid the study of camera-based rPPG algorithms for HR and HRV analysis.

Ground truth signals. Another significant but overlooked complication with existing rPPG datasets arises from their provided ground truth signals, typically photoplethysmogram (PPG) or electrocardiogram (ECG) waveform. These signals are often plagued by artefacts, for example in the form of large spikes caused by unwanted sensor movement and interference [1,14]. This causes false peak detections resulting in incorrect ground truth HR and HRV measures, thereby leading to unreliable evaluation. Additionally, directly utilizing raw PPG/ECG signals as ground truth leads to another issue in the evaluation process: evaluatees are free to choose the method of obtaining peaks on them, and they could use the very same peak detection algorithm that is used by the method under evaluation on the rPPG signal. Since the estimated rPPG peaks are evaluated against ground truth peaks generated by the same algorithm, this can lead to a "detector-bias" in the computed error measures. For example, an algorithm that blindly detects a fixed number of heart-beats on all signals will incorrectly report zero error.

To help solve these problems caused by noisy ground truth signals, we introduce CleanerPPG: a public collection of human-verified peak/heart-beat annotations on ground truth signals of existing publicly available rPPG datasets. This makes the ground truth heart-beats absolute and independent, and therefore offers more accurate, fairer and standardized evaluations and benchmarking. CleanerPPG is made publicly available* and intended to develop into a continuously growing community-driven collection for all future datasets.

Contributions. We make the following contributions in this work:

(i) We present an efficient unsupervised rPPG pipeline that can estimate heart-rate from RGB webcams. This method has the advantage that it does not require any rPPG specific training and it can perform its analysis with real-time speeds.
(ii) Our method is able to time individual heart beats in the estimated pulse signal to compute heart rate variability. This body of work has received little attention, and we set the first benchmarks on multiple public datasets.
(iii) We perform an in-depth HR and HRV evaluation on an exhaustive collection of 13 public and self-recorded datasets exploring a varied range of unique facets. We also demonstrate state-of-the-art level performance on six public datasets.
(iv) We introduce a new publicly available high frame-rate dataset, VicarPPG 2, specifically designed to evaluate rPPG algorithms under various subject conditions for HR and HRV analysis.
(v) Lastly, we tackle the problem of noisy ground truth signals and the peak detector bias by releasing a collection of hand-cleaned heart-beat peaks for existing public datasets.

## 2. Related Work

Signal processing based rPPG methods. Since the early work of Verkruysse *et al.*[3], who showed that heart rate could be measured from consumer grade camera recordings in ambient light, a large body of research has been conducted on the topic. Extensive reviews of these work can be found in [2,13,15]. Most published rPPG methods work either by applying skin detection on a certain area in each frame or by selecting one or multiple regions of interest and track their averages over time to generate colour signals. A general division can be made into methods that use blind source separation (ICA, PCA) [16–18] vs those that use a 'fixed' extraction scheme for obtaining the blood volume pulse (BVP) signal [19–23]. The blind source separation methods require an additional selection step to extract the most informative BVP component signal. To avoid this, we make use of a 'fixed' extraction scheme in our method.

---

* This dataset is available for research purposes, and can be requested via the link: www.vicarvision.nl/datasets/vicarppg2

Among the 'fixed' methods, multiple stand out and serve as inspiration and foundation for this work. Tasli *et al.* [21] presented the first face modelling based signal extraction method and utilized detrending [24] based filtering to estimate BVP and heart rate. The CHROM [19] method uses a ratio of chrominance signals which are obtained from RGB channels followed by a skin-tone standardization step. Li *et al.* [20] proposed an extra illumination rectification step using the colour of the background to counter illumination variations. The SAMC [22] method proposes an approach for BVP extraction in which regions of interest are dynamically chosen using self adaptive matrix completion. The Plane-orthogonal to skin (POS) [23] method improves on CHROM. It works by projecting RGB signals on a plane orthogonal to a normalized skin tone in normalized RGB space, and combines the resulting signals into a single signal containing the photopleytismographic information. We take inspiration from Tasli *et al.* [21] and further build upon POS [23]. We introduce additional signal refinement steps for accurate peak detection to further improve HR and HRV analysis.

Deep learning based rPPG methods.　Most recent works have applied deep learning to extract either heart rate or the blood volume pulse directly from camera images. They rely on the ability of deep networks to *learn* which areas in the image correspond to heart rate. This way, no prior domain knowledge is needed and the system learns the underlying rPPG mechanism from scratch. DeepPhys [5] is the first such end-to-end method to extract heart and breathing rate from videos. HR-CNN [6] uses two successive convolutional neural networks (CNNs) [25] to first extract a BVP from a sequence of images and then estimate the heart rate from it. RhythmNet [26] uses a CNN and gated recurrent units to form a spatiotemporal representation for HR estimation. The recent work of AutoHR [27] employs neural architecture search to discover temporal difference convolution as a strong backbone to capture the rPPG signal from frame sequences. These methods have shown state-of-the-art performance on multiple public and private datasets. Our presented algorithm is unsupervised and makes use of an active appearance model [28] to select regions of interest to extract a heart rate signal from. Due to this, no rPPG specific model training is required while prior domain knowledge is more heavily relied upon.

Heart Rate Variability from rPPG.　Some past methods have also attempted extracting heart rate variability from videos [18,29,30]. A good overview of this is provided by Rodriguez *et al.* [31]. Because HRV is calculated based on variations in inter-beat intervals, it is crucial that single beats are detected and localized with a high degree of accuracy. Methods that otherwise show good performance in extracting HR can be unsuitable for HRV analysis since they may not provide beat locations. Rodriguez *et al.* [31] evaluate their baseline rPPG method for HRV estimation. Their method is based on bandpass filtering the green channel from regions of interest. However, their results are only reported on their own private dataset (not publicly available), which makes direct comparison difficult. More recent works have shown and benchmarked video-based HRV measurement on publicly available datasets. Finžgar and Podržaj [32] introduce a wavelet transform and custom inter-beat-interval filtering rPPG algorithm and evaluate it on the publicly available PURE dataset [33]. They shown good correlation between time-domain ultra-short term HRV measurements from rPPG and PPG. Work by Li *et al.* [34] highlight the effectiveness of a clear signal for peak detection, and apply a slope sum function to create more pronounced peaks in the rPPG signal. Concurrent work by Song *et al.* [35] introduces one of the first deep learning based fully supervised techniques for HRV estimation, relying on a generative adversarial network to learn denoising of rPPG signals. Both these papers report their results on the UBFC-RPPG dataset [36], which is publicly available. Our method also estimates heart rate variability by obtaining precise temporal beat locations from the filtered BVP/rPPG signal, and we report our HRV results on a large number of public datasets.

rPPG datasets.　Due to a scarcity of rPPG datasets in the past, initial attempts at evaluating rPPG methods were on private self-recorded videos [23,29,37]. Some of the earliest publicly available datasets with heart rate annotations repurposed for rPPG research were introduced in [11] and [12], both of which were originally recorded for the purpose of psychological studies. Although the lab setting and video compression makes some of these datasets less than ideal for rPPG, their public availability and large sample sizes provide a common platform for benchmarking rPPG methods. More recently, [21,33,38,39] introduced datasets specifically recorded with the intention of being used for remote heart rate estimation research. These sets include variations in illumination, physical/physiological conditions, and camera types. In this paper, we introduce a new high frame-rate video dataset for rPPG evaluation designed with a focus on evaluating short-term heart rate variability estimation, which require longer observations. This dataset is made publicly available for research use[†].

## 3. Method

We present a method for extracting heart rate (HR) and heart rate variability (HRV) from the face in real-time using only a consumer grade webcam and CPU, as shown in Figure 1.

### 3.1. Skin pixel selection

The first step in the pipeline includes face finding [40] and fitting an active appearance model (AAM) [28]. This AAM is then used to determine facial landmarks (from the AAM shape vector) as well as the head orientation (by measuring angular deviation from the mean frontal pose). The landmarks are used to define a region of interest (RoI) which only contains pixels on the face belonging to skin. This allows us to robustly track the pixels in this RoI over the course of the whole video. Our RoI consists of the upper region of the face excluding the eyes (determined empirically). An illustration of this can be seen in Figure 1. The head orientation is used to measure and track the pitch, roll, and yaw angles of the head per frame. Across all pixels in the RoI, the averages for each colour channel (R,G,B) is computed and tracked (concatenated) to create three colour signals.

### 3.2. Signal extraction

The colour signals and the head orientation angles are tracked over a running time window of 8.53 seconds, which corresponds to 256 frames at 30 fps, or 512 frames at 60 fps. To counteract the impact of variations in frame rates of the input, all signals are resampled (using linear interpolation) to a fixed sampling rate of 30 or 60 Hz, whichever is closer to the frame rate of the source video. The choice of this window duration and sampling rate is based on the resulting signal length being a power of two, which is compatible with optimized fast Fourier transform operations. Subsequently, the three colour signals from R, G and B channels are combined into a single rPPG signal using the POS method [23]. This method filters out intensity variations by projecting the R, G and B signals on a plane orthogonal to an empirically determined normalized skin tone vector. The resulting 2-D signal is combined into a 1-D signal via a weighted sum with the weight determined by the ratio of standard deviations of the two signals. This ensures that the resulting rPPG signal contains the maximum amount of the pulsating component.

### 3.3. Signal filtering

#### 3.3.1. Rhythmic motion noise suppression

A copy of the extracted rPPG signal as well as the head-orientation signals are converted to the frequency domain using Fast Fourier Transform. The three resulting head-orientation spectra (one each of pitch, roll, and yaw) are combined into one via averaging. This is then subtracted from the raw rPPG spectrum after amplitude normalization. This way, the frequency components having a high value in the head-orientation spectrum are attenuated in the rPPG spectrum. Subsequently, the frequencies outside of the human heart rate range (0.7 - 4 Hz / 42 - 240 bpm) are removed from the spectra.

#### 3.3.2. Wide & narrow band filtering

The strongest frequency component inside the resulting spectrum is then used to determine the passband range of a narrow-bandpass filter with a bandwidth of 0.47 Hz. This bandwidth has been chosen empirically and depends on the robustness of the subsequent peak detection algorithm to distinguish heat beat peaks from noise (higher the robustness, wider this bandwidth can be). This bandpass filter can either be realized via inverse FFT or a high order FIR filter (e.g. ~50$^{th}$ order Butterworth). The selected filter is then applied to the original extracted rPPG signal to produce noise-free BVP.

---

[†] This dataset can be requested via the link: www.vicarvision.nl/datasets/cleanerppg

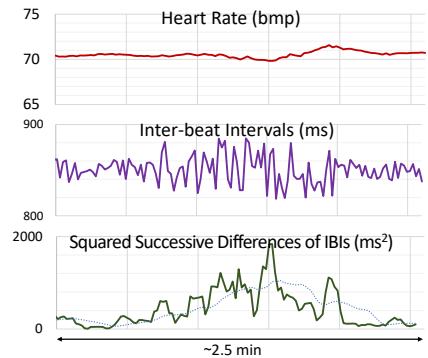

**Figure 2.** Example of heart rate variability computation: Even when the heart rate (HR) is almost constant, the underlying inter-beat intervals (IBIs) can have many fluctuations. This is detected by rising squared successive differences (SSD), a measure of heart rate variability.

*3.4. Post processing*

To prevent minor shifts in the locations of the crest of each beat over multiple overlapping running windows, the signals from each window are overlap added with earlier signals [19,23,41]. First, the filtered rPPG signal is normalized by subtracting its mean and dividing it by its standard deviation. During resampling of the signal, the number of samples to shift is determined based on the source and resampled frame rates. The signal is then shifted back in time accordingly and added to the previous/already overlapped signals. Older values are divided by the times they have been overlap added, to ensure all temporal locations lie in the same amplitude range. Over time, a cleaner rPPG signal is obtained from this.

*3.5. Output calculation*

Once a clean rPPG signal is obtained, we can perform peak detection on it to locate the individual beats in time in the signal. From the located beats, heart rate and heart rate variability can be calculated. To do this, we first extract the inter-beat-intervals (IBIs) from the signal, which are the time intervals between consecutive beats.

3.5.1. Inter-beat interval pre-filtering

Before calculating HR/HRV, the extracted inter-beat intervals (IBI) are filtered to remove noise caused by false positive/negative peak detections. First, all IBIs lying outside the range of 250 ms to 2000 ms are excluded (corresponding to the human heart rate range of 30 to 240 bpm). To further remove strong outliers from the signals, intervals farther than three standard deviations from the mean are removed.

3.5.2. Heart rate calculation

Heart rate is calculated by averaging all IBIs over a time window, and computing the inverse of it. That is, $HR_w = 1/\overline{IBI}_w$, where $\overline{IBI}_w$ is the mean of all inter-beat intervals that fall within the time window $w$. This gives the heart rate in Hertz (assuming IBIs in seconds), and multiplying by 60 gives us the heart rate in beats-per-minute. The choice of this time window can be based on the user's requirement (e.g. instantaneous HR, long-term HR).

3.5.3. Heart rate variability calculation

Multiple metrics can be computed to express the measure of heart rate variability in different units. In this work, we focus on one of the most commonly used time-domain metric for summarizing HRV called the 'root mean square of successive differences' (RMSSD) [10,31,37,42], expressed in units of time. As the name suggests, this is computed by calculating the root mean square of time difference between adjacent IBIs:

| PURE [33] | UBFC-RPPG [36] | MMSE-HR [45] | VIPL-HR [38] | ECG-Fitness [6] |
|---|---|---|---|---|
| 10 subjects / 59 videos. Subjects recorded with varying movement patterns: talking, slow/fast translation, and small/large rotation. 480p @ 30fps. PNG images [lossless]. Ground truth PPG @ 60Hz. | 42 subjects / 42 videos. Subjects recorded while playing a stressful game (under "realistic" partition). 480p @~30fps. Raw video format [lossless]. Ground truth PPG @ 30/60Hz. | 40 subjects / 102 videos. Part of a larger multi-modal corpus containing recordings while subjects exhibit facial expressions. 1040x1392p @25fps. JPEG images. Instantaneous HR @ 1KHz. | 107 subjects / 2378 videos. Large dataset with a range of movement, illumination, and camera types. ~460x502p face crops. 25~30fps – MJPEG format. Ground truth PPG @ 60Hz. | 17 subjects / 17 videos. Recordings while exercising on rower, bike and elliptical equipment, and also while talking. 1080p @ 30fps. YUV format [lossless]. Ground truth ECG @ ~125Hz. |
| **MAHNOB-HCI [11]** | **VicarPPG [21]** | **DEAP [12]** | **MoLi-PPG [39]** | **COHFACE [46]** |
| 27 subjects / 527 videos. Subjects recorded while watching video stimuli. 780×580p @ 61fps. H.264 format. Ground truth ECG @ 256Hz. | 10 subjects / 20 videos. Unrestrained subjects recorded before and after performing strenuous workout. 720p @~30fps [variable]. H.264 format. Ground truth PPG @ 30Hz. | 874 videos of 22 subjects. Subjects recorded while watching music videos. Faces are significantly occluded by electrodes. 720x576p @ 50fps. H.264 format. Ground truth PPG @ 128Hz. | 170 videos of 30 subjects. Subjects recorded under varying illumination, movement, and speech. 1080p/720p/600p @ 25/50fps. MPEG-4 Part 2 format. Ground truth ECG @ 256Hz. | 164 videos of 40 subjects. Subjects recorded illuminated by a spotlight and by uneven natural light. 480p @ 20fps. MPEG-4 Part 2 format. Ground truth PPG @ 256Hz. |

**Figure 3.** A list of the previously available rPPG datasets used in the this paper, along with their key details.

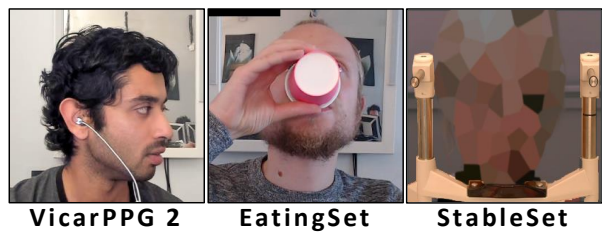

**Figure 4.** Example images from newly introduced datasets: (left to right) VicarPPG 2, EatingSet, and StableSet. The example from VicarPPG 2 shows subject suddenly turning their head. The EatingSet image shows subject taking a long sip resulting in face occlusion. The subjects in StableSet were physically stabilized using the shown chin rest (face removed for privacy reasons).

$$\text{RMSSD} = \sqrt{\frac{1}{N}\sum_{i=1}^{N}(\text{IBI}_i - \text{IBI}_{i+1})^2}, \tag{1}$$

where $\text{IBI}_i$ represents the $i^{\text{th}}$ inter-beat interval, and $N$ represents the number of IBIs in the sequence. A graphical example of such HRV calculation is shown in Figure 2. Because RMSSD is more susceptible to noise, only IBIs within the first standard deviation around the mean are considered. Along with RMSSD, we calculate another time-domain HRV metric known as the 'standard deviation of NN intervals' (SDNN) [10], which is simply the standard deviation of all filtered IBIs in the sequence.

In addition, we also compute two frequency-domain metrics of HRV, simply known as 'low-frequency' (LF) and 'high-frequency' (HF) bands [42] (as well as a ratio of them), that are commonly used in rPPG HRV literature [16,29,31]. The LF and HF components are calculated using Welch's power spectral density estimation [43]. Since Welch's method expects evenly sampled data, the IBIs are interpolated at a frequency of 4Hz using spline interpolation and detrended to remove very low frequency components [24]. The power of each band is calculated as total power in a region of the periodogram: the LF band from [0.04 to 0.15 Hz], and the HF band from [0.15 to 0.4 Hz]. Both metrics are converted to normalized units by dividing them by the sum of LF and HF. Details about these metrics can be found in [10,44].

## 4. Datasets

To compare against prior work and study the properties of the proposed method, we evaluate on a comprehensive collection of datasets. Table 3 provides summarized details of existing datasets used in this paper, while the self-recorded ones are listed below. Some example frames from these self-recorded datasets are shown in Figure 4.

<u>StableSet rPPG dataset.</u> To make a proof-of-concept test of our proposed rPPG method, the StableSet rPPG dataset was collected. This dataset contains recordings of participants while they were shown short video clip stimuli on a screen facing them. During the recording of this dataset, the participating subjects' head movements were physically stabilized using a chin rest with the intention of minimizing motion induced noise in rPPG measurements. A total of 24 participants were included in this dataset, aged between 18 and 30 years (with a mean of 21.5 years), and having a male : female gender ratio of 9 : 15. Ground Truth was collected in the form of ECG (via a Mobi8 device) at a sampling rate of 1 KHz, and the videos were recorded using a front-facing HD camcorder (JVC GZ-VX815) with a resolution of 1920×1080 pixels at a framerate of 25 fps. Camera settings like brightness, aperture, backlight compensation and white balance were set to manual on the camcorder to keep filming conditions ideal and constant. Prior informed consent was obtained from all participants. However, due to privacy restrictions within the consent, the videos of this dataset are not made available publicly.

<u>VicarPPG-2 dataset.</u> Specifically aimed at evaluating rPPG algorithms at estimating heart rate and short-term heart rate variability (which requires a minimum observation of 5 minutes [10]), we recorded the VicarPPG-2 Dataset. 10 subjects participated in the data collection. The male : female ratio was 7 : 3 with an average age of $29 \pm 5$ years, and skin types ranging on a Fitzpatrick scale [47] from II to IV.

Participants were asked to sit in front of a computer screen (∼1 meter distance) on which the instructions were shown, a webcam was mounted on top of the screen and an LED ring lamp was mounted behind the camera. Screen brightness was reduced as far as possible to minimize the influence of screen light on the face. All videos were recorded using a Logitech Brio webcam at a fixed framerate of 60 fps using an H.264 compliant encoder (Microsoft Media Foundation), and stored in `mp4` containers. The recording location was illuminated by natural ambient light in addition to a LED ring lamp (Falcon Eyes DVR-300DVC) to prevent strong shadows and influences of large changes in natural light. The ground truth signals were recoded in the form of synchronized ECG signals at 250 Hz sampling frequency obtained via an Arduino based ECG board (AD8232), and synchronized PPG signals at 60 Hz obtained via a pulse oximeter device (CMS50E) attached to the left index finger of the participant. The following four scenarios/conditions were recorded for each participant:

(**i**) Baseline: Participants sitting naturally while watching a relaxing video or reading an article on screen.
(**ii**) Movement: participants performing four different types of pre-planned angular body/head movements: turning head side-to-side (shaking), moving head up and down (nodding), a combination of head shaking and nodding (round), moving eyes while keeping head still, and naturally bobbing their heads while listening to music (dance).
(**iii**) Stress: Participants playing a stress-inducing Stroop effect [48] based game.
(**iv**) Post-workout: Participants sitting unrestrained after performing fatigue-inducing physical workouts to induce higher heart rates.

Each condition was recorded for a duration of 5 minutes to allow for the computation of short-term heart rate variabilty, a total of 200 minutes of video were collected. Out of 400 collected ground truth files, two had to be removed from the dataset due to excessive finger movement in the PPG device and gradual detachment of the ECG ground electrode, leading to unusable signals. This dataset stands out as it was explicitly collected for RPPG purposes featuring 5-minute long 60 fps camera recordings under various physical/pysiological conditions, with simultaneous ECG and PPG ground truth recordings. Informed consent was obtained from all participating subjects. This dataset is available for research purpose, and can be requested via the link: www.vicarvision.nl/datasets/vicarppg2.

<u>EatingSet rPPG dataset.</u> This is a self-recorded video dataset comprising of 20 subjects, with a male : female ratio of 14 : 6. The average age in the dataset was $32 \pm 8.6$ years, and the subjects had skin types ranging on the Fitzpatrick scale [47] from II to IV. The recording setup and conditions were similar to those of VicarPPG 2. Participants were sitting at 1 meter distance to a screen on which instructions were shown, while being illuminated by an LED ring lamp. All videos were recorded using a Logitech C920 webcam at 30 fps in uncompressed YUYV422 pixel format. PPG signals were collected as ground truth, at a frequency of 60 Hz, via a pulse oximeter device (CMS50E) attached to the left index finger of the participant. During recording participants were asked to consume 4 types of food items with varying consistency. These include a sip of water (drink), a cookie (crumbly), a marshmallow (chewy) and multiple almonds (hard). Informed consent was obtained from all participants. However, due to privacy restrictions, the videos of this dataset are not made available publicly. This unique dataset contains a variety of natural, non-rhythmic deformations and facial occlusions while eating as they might occur in real-world situations, and serves as a challenging testbed for rPPG evaluation.

CleanerPPG ground truth dataset. Finally, we introduce a meta-dataset which contains cleaned ground truth signals of already existing datasets. All peaks/beats detected from ground truth signals (ECG, PPG) of all datasets used in this paper were hand-verified and corrected by human expert annotators, in order to achieve a more accurate and standardized evaluation.

To do this, candidate peaks were first obtained using a gradient-based signal peak detector [49]. Following this, a human verification step was performed wherein an expert matched, verified and corrected every candidate peak by observing the shape of the raw PPG/ECG waveform and the inter-beat intervals. A specialized Python tool was developed for this step, which allow annotation of single peaks as well as zooming and an overview of the resulting RR intervals. Peaks were annotated at the crest of the PPG signal or at the highest point of the R peak in the ECG signal. Ectopic beats (genuine extra heart beats that can occur between two regular beats) in the ECG signal were included for the sake of completion. Parts of the signal that were too noisy due to movement of electrodes or finger to make out clear peaks were left blank.

These steps led to the removal of false positive and negative peak detections caused by artefacts in the signal, resulting in a collection of noise-free ground truth heart beat annotations for an exhaustive set of publicly available rPPG datasets. Annotators spent an average of ~30 seconds per minute of signal duration, results in a total of 36 person-hours of annotation work for cleaning 75 hours of ground truth data. These annotations can be used to perform a more accurate evaluation of rPPG methods, especially for noise-sensitive measures like HRV. These peak annotations can also be used for training fully supervised machine learning methods to be able to distinguish between true heat beat peaks and noise, thereby improving accuracy of such methods. This collection is available for research purposes, and can be requested via the link: www.vicarvision.nl/datasets/cleanerppg. All experiments performed in this paper utilize these hand-cleaned peaks.

## 5. Experiments and Results

### 5.1. Impact of ground truth peak cleaning: CleanerPPG

To lay foundation for the rest of the experiments in this paper, we first study the impact and value of evaluating rPPG methods against hand-cleaned ground truth. Figure 5 provides examples of the kind of artefacts that plague raw ground truth PPG/ECG signals resulting in incorrect peak detections. To provide a more qualitative analysis, we measure the agreement between the raw ground truth peaks (obtained via a gradient-based peak detector [49]) and the hand-cleaned ground truth peaks from CleanerPPG. We do this by computing the mean absolute error/difference between their computed HR and HRV values on all datasets. The results of this study can be seen in Table 1.

On average over all datasets, the heart rate values computed from the raw ground truth peaks deviate from the clean peaks in the range of 0.01 to 4.3 bpm, with almost all datasets containing videos that deviate over 10 bpm. For some datasets (VicarPPG {1&2}, UBFC, MAHNOB, PURE, COHFACE, VIPL, DEAP), the raw PPG/ECG signals provided are already of fairly good quality with few artefacts, resulting in a relatively small average deviation from the cleaned peaks (less than ~0.5 bpm). On the other hand, many datasets exhibit significant deviations from the cleaned ground truth (MoLi-PPG, StableSet), with some diverging quite heavily (EatingSet, MMSE-HR, ECG-Fitness). Similar but larger deviations are also observed for HRV (RMSSD) calculations, which is much more sensitive to fine errors in peak detection.

These deviations can be problematic when comparing/benchmarking the performance of rPPG methods against each other. This is especially the case when the gap in performance between methods is small. For example, the raw ground truth of PURE deviates from cleaned ones by 0.36 bpm. While small, this is significant since the top two state-of-the-art methods have a performance gap of only 0.3 bpm on this dataset (as seen in Table 2).

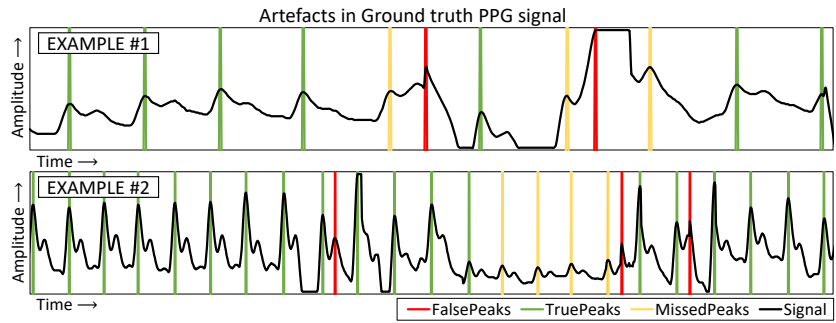

**Figure 5.** Examples of peak detection on artefact-prone raw ground truth PPG signals. Spikes in the amplitude result in false peak detections while some regions of the signal are attenuated resulting in true peaks not being detected.

| Dataset | HR (bpm) | HRV [RMSSD] (ms) | Dataset | HR (bpm) | HRV [RMSSD] (ms) |
|---|---|---|---|---|---|
| PURE | 0.34 ±1.8 | 7.9 ±13.1 | MoLi-PPG | 1.6 ±9.3 | 4.4 ±16.1 |
| UBFC-RPPG | 0.24 ±0.6 | 11.2 ±8.4 | DEAP | 0.41 ±1.7 | 11.85 ±37.4 |
| MMSE-HR | 3.57 ±3.5 | — | VicarPPG | 0.01 ±0.02 | 0.11 ±0.3 |
| VIPL-HR | 0.57 ±3 | 17.33 ±57.9 | VicarPPG 2 | 0.08 ±0.5 | 4.52 ±19.4 |
| COHFACE | 0.46 ±2.2 | 14.36 ±49.6 | EatingSet | 4.33 ±1.6 | 6.54 ±10.1 |
| ECG-Fitness | 3.47 ±15.2 | 19.24 ±76.1 | StableSet | 1.11 ±6.1 | 10.153 ±43.6 |
| | | | MAHNOB | 0.27 ±1.8 | 3.31 ±23.0 |

**Table 1.** Deviations (in terms of mean absolute error) in HR (bpm) and HRV [RMSSD] (ms) calculations between the raw ground truth and the hand-cleaned peaks from the CleanerPPG set. Many datasets exhibit notable deviations due to artefacts in the raw signal. This leads to miscalculation of error metrics used for evaluation.

These results suggest that using the raw ground truth peaks for evaluation can result in substantial miscalculations of error metrics, potentially leading to incorrect conclusions. In all the experiments that follow, the cleaned CleanerPPG peaks are used as the ground truth.

## 5.2. Benchmarking and comparison against state-of-the-art

In order to study the generalizability of the proposed rPPG method, we benchmark it's performance on a range of datasets. To assess accuracy, we measure the deviation of the predicted HR/HRV measures from the ground truth in terms of mean absolute error (MAE), which is the average of the absolute differences between predicted and true values (obtained within a set time window for HR). The results for heart rate and heart rate variability analysis are listed in Tables 2 and 3 respectively. These results are also compared against prior work to provide context w.r.t. the state of the art. In addition, results of a blind baseline estimator that always predicts a heart rate of 75 bpm (mean HR of all datasets) are also included in the table for additional context.

### 5.2.1. Heart Rate Estimation

The results for heart rate analysis are listed in Tables 2. Here, while comparing, it should be noted that the supervised methods train or fine-tune on part of the dataset they are evaluated on, while unsupervised methods do not. Further note that the results for most supervised methods and some unsupervised methods (marked with a *) are reported on a smaller unspecified subset (test set) of the dataset. Also, the reported heart rate mean absolute errors are computed using different time-window sizes (denoted within the brackets $^{\{\}}$ when known). To aid comparison, we report our results on three most commonly used window sizes: 15 secs, 30 secs and full-video length (denoted by $^{\{\infty\}}$).

The proposed method performs well at the state-of-the-art level or beyond on most of the datasets: PURE, VIPL-HR, MoLiPPG, VicarPPG, UBFC-RPPG, MMSE-HR. The performance on these datasets is not only better than other unsupervised signal-processing methods, but also better or on par with fully supervised deep learning methods. The very high accuracy of 0.22 - 0.36 bpm on the PURE dataset can be attributed to the videos being stored in a lossless format and thus having no compression noise. The low error rate on StableSet can be attributed to the fact that subjects' movements were physically stabilized via a chin-rest (see Figure 4). On MMSE-HR, our method is able to perform well despite the subjects showing a range of facial expressions (e.g. laughter). This can be attributed to the robustness of the face modelling step. Conversely on DEAP, the accuracy of the algorithm is moderate, with the likely source of errors being poor face modelling due to the presence of electrodes occluding the face. On the ECG-Fitness dataset, although the proposed methods performs better or similar to other unsupervised methods, the accuracy is quite poor. This is largely caused by the extremely high intensity movements of the subjects while performing physical exercises.

| Method | | PURE | UBFC-RPPG | MMSE-HR | VIPL-HR | COHFACE | ECG-FITNESS | MAHNOB-HCI | MoLi-PPG | DEAP | VicarPPG | VicarPPG 2 | EatingSet | StableSet |
|---|---|---|---|---|---|---|---|---|---|---|---|---|---|---|
| | Baseline (75 bpm) | 17.6 ±15.1 | 13.7 ±9.1 | 12.6 ±12.4 | 9.16 ±8.5 | 10.7 ±5.8 | 35.3 ±21.6 | 9.71 ±6.6 | 8.98 ±8.3 | 8.03 ±5.7 | 16.2 ±10.8 | 8.38 ±6.1 | 10.1 ±7.2 | 7.9 ±6.1 |
| | FaceRPPG [Ours] | 0.36 ±0.4 {15} | 2.6 ±6.7 {15} | 3.82 ±10.1 {15} | 10.6 ±11.5 {15} | 12 ±9.3 {15} | 24.9 ±22.9 {15} | 16.9 ±13.5 {15} | 4.85 ±7.2 {15} | 7.17 ±5.9 {15} | 1.83 ±3.5 {15} | 3.43 ±5.1 {15} | 3.18 ±2.4 {15} | 0.9 ±1.6 {15} |
| | | 0.26 ±0.3 {30} | 2.37 ±7.1 {30} | 3.73 ±10.1 {30} | 10 ±11 {30} | 10.8 ±9.8 {30} | 24.9 ±22.4 {30} | 13.1 ±10.4 {30} | 4.45 ±7.1 {30} | 6.24 ±6.1 {30} | 1.28 ±2.2 {30} | 3.09 ±5.0 {30} | 2.97 ±2.5 {30} | 0.7 ±1.6 {30} |
| | | 0.22 ±0.2 {∞} | 1.98 ±6.2 {∞} | 3.89 ±10.3 {∞} | 9.88 ±10.9 {∞} | 10.4 ±9.5 {∞} | 24.8 ±22 {∞} | 13.1 ±10.4 {∞} | 3.5 ±6.9 {∞} | 5.57 ±5.5 {∞} | 1.53 ±3.5 {∞} | 2.59 ±4.3 {∞} | 2.77 ±2.7 {∞} | 0.59 ±1.6 {∞} |
| Unsupervised Signal Processing Methods | Green/EVM[7] | | | | | | | | | | 5.6 {15} [21] | | | |
| | Tasli et al. [21] | | | | | | | | | | 4.2 {15} | | | |
| | C-MCCA[50] | | | | | | | | | 3.66 | | | | |
| | cPR+fine[39] | | 2.1 | | | | | | 6.13 | | | | | |
| | POS[23] | | 4.73 {30} [51] | 5.77 {15} [51] | 11.5 {10} [26] | | | | | | | | | |
| | ICA[52] | *24.1 [53] | 6.02 {30} [51] | 5.84 {15} [51] | | | | | | | | | | |
| | NMD-HR[53] | *8.68 | | | | | | | | | | | | |
| | 2SR[54] | *2.44 {∞} | | | | *20.98 {∞} [6] | *43.66 {∞} [6] | *13.84 {∞} [6] | | | | | | |
| | CHROM[19] | *2.07 {∞} | 3.7 {30} [51] | 5.59 {15} [51] | ≤16.9 {10} [26] | *7.8 {∞} [6] | *21.37 {∞} [6] | *13.49 {∞} [6] | | | | | | |
| | LiCVPR[20] | *28.2 {∞} | | ≤19.95 {10} [26] | | *19.98 {∞} [6] | *63.25 {∞} [6] | *7.41 {∞} [6] | | | | | | |
| | Wavelet[55] | | | 2.4 {10} | | | | | | | | | | |
| | PVM[51] | | 4.47 {30} | 4.38 {15} | | | | | | | | | | |
| | MAICA[56] | | 3.43 | 3.91 | | | | | | | | | | |
| | VD+LMS+HRE[57] | | | | 25.52 | | | | | | | | | |
| | CK[58] | | 2.29 | | | | | | | | | | | |
| Supervised ML Methods | HR-CNN[6] | *1.84 {∞} | | | | *8.1 {∞} | *14.48 {∞} | *7.25 {∞} | | | | | | |
| | SAMC[22] | | | ≤*11.37 | 15.9 {10} [26] | ≤6.23 | | | | | | | | |
| | ST-CNN[59] | *0.87 | | | | | | | | | | | | |
| | Attentn-CNN[60] | | | | | | | *6.8 | | | | | | |
| | DeepPhys[5] | | | | *11 {10} [26] | | | 4.57 {∞} | | | | | | |
| | RythmNet[26] | | | ≤*5.03 {10} | *5.3 {10} | | | ≤*4.00 {∞} | | | | | | |
| | Siamese-rPPG[61] | *0.63 | *0.48 | | | | | | | | | | | |
| | AutoHR[27] | | | 5.87 | *5.68 | | | *3.78 {∞} | | | | | | |
| | PhysNet[62] | | | ≤13.25 [27] | *10.8 [27] | | | 5.96 {∞} | | | | | | |
| | PulseGAN[35] | | 2.09 | | | | | | | | | | | |

**Table 2.** A comparison of the performances of various methods in terms the mean absolute error in beats per minute (bpm). {15}, {30}, and {∞} represent the HR calculation windows of 15 s, 30 s, and full-video length respectively. * represent accuracies obtained on a smaller (test) subset of the full dataset. ≤ represents root mean squared error, which is always greater than or equal to mean absolute error. Baseline represents the accuracy obtained by always predicting a heart rate of 75 bpm (average HR over all datasets). The reported results of the proposed FaceRPPG method (and the baseline) are against the cleaned ground truth from the CleanerPPG set. The references next to the results denote the source from which they were obtained. The different colours separate the methods into two different categories: unsupervised signal processing methods, and fully supervised deep learning methods. Note that the supervised methods also train or fine-tune their parameters on parts of dataset they are being evaluated on, while unsupervised methods do not. The proposed method outperforms most prior work, including fully supervised ones.

| Method | HRV Metric | PURE | UBFC-RPPG | VIPL-HR | COHFACE | ECG-FITNESS | MAHNOB-HCI | MoLi-PPG | DEAP | VicarPPG | VicarPPG 2 | EatingSet | StableSet |
|---|---|---|---|---|---|---|---|---|---|---|---|---|---|
| FaceRPPG [Ours] | RMSSD (ms) | 15 ±12.7 | 16.3 ±22.5 | 72.7 ±57.8 | 119 ±44.4 | 81.9 ±53.2 | 108 ±51.4 | 42.9 ±43.7 | 74.4 ±41.4 | 21.9 ±13.8 | 26.5 ±18.9 | 37.1 ±33 | 21 ±37.9 |
| | SDNN (ms) | 18.4 ±10.4 | 18.7 ±14.9 | 49.4 ±45.5 | 79.8 ±39.6 | 53.2 ±48.2 | 107 ±51.8 | 35.6 ±30.2 | 46.2 ±35.2 | 44.1 ±28.1 | 16.4 ±15.3 | 15.8 ±11.5 | 21.8 ±23.8 |
| | LF & HF (n.u.) | 0.13 ±0.1 | 0.17 ±0.13 | 0.25 ±0.18 | 0.24 ±0.17 | 0.3 ±0.2 | 0.3 ±0.2 | 0.17 ±0.13 | 0.28 ±0.18 | 0.13 ±0.1 | 0.2 ±0.11 | 0.28 ±0.18 | 0.13 ±0.1 |
| | LF/HF | 1.27 ±3.03 | 1.0 ±0.99 | 1.61 ±2.8 | 2.35 ±8.01 | 3.18 ±4.37 | 2.89 ±11.8 | 0.99 ±1.22 | 1.88 ±2.76 | 0.53 ±0.36 | 1.53 ±1.43 | 1.54 ±1.44 | 0.54 ±0.76 |
| Finžgar et. al. [32] | RMSSD (ms) | 16.77 | | | | | | | | | | | |
| | SDNN (ms) | 8.14 | | | | | | | | | | | |
| SSF[34] | RMSSD (ms) | | 47 | | | | | | | | | | |
| | SDNN (ms) | | 25 | | | | | | | | | | |
| CHROM[19] | RMSSD (ms) | | 93 [34] | | | | | | | | | | |
| | SDNN (ms) | | 38.9[35] | | | | | | | | | | |
| PulseGAN[35] | SDNN (ms) | | 24.3 | | | | | | | | | | |

**Table 3.** Heart rate variability estimation performance in terms of mean absolute error. The metrics included are RMSSD and SDNN (in milliseconds), LF and HF (in normalized units), and the ratio of LF/HF. The different colours separate the different metric types as denoted in the second column. The proposed FaceRPPG method outperforms all prior work, inlcuding fully supervised ones. It performs well on datasets with low video compression noise or limited subject movement, but fails when these factors become large. FaceRPPG results are reported against the cleaned ground truth from the CleanerPPG set.

On VIPL, the source of a majority of the errors are the videos recorded on a mobile device. This could be because of relatively inferior camera sensor and/or stronger compression on such a device. On the videos of MAHNOB-HCI, which are also highly-compressed, we see that our method does not achieve a very good accuracy, similar to the majority of unsupervised signal processing methods. An interesting observation is that the error produced by almost all unsupervised methods is higher than that of a dummy baseline method that blindly predicts a heart rate of 75 bpm for any input (this is also the case for VIPL and COHFACE). Only the supervised methods are able to perform better (direct comparison not always possible since fine-tuning is performed and the reported results are on an unspecified test subset). This suggests that the high compression noise distorts the pulse information in the spatial averages of skin pixels. Deep learning based methods seem to be able to somewhat overcome this, perhaps by learning to detect and filter out the spatial 'pattern' of such compression noise.

5.2.2. Heart Rate Variability Estimation

The task of assessing HRV is much more noise-sensitive than estimating heart rate. In Table 3, the results of heart rate variability estimation are listed for all datasets. Since HRV is a relatively long-term measure, these HRV metrics are computed over complete video lengths. Our unsupervised method sets the first HRV evaluation benchmarks on most of the datasets, and outperforms all previous methods on PURE and UBFC-RPPG datasets, including deep learning based fully supervised ones (*PulseGAN* [35]).

Based on HRV literature [10] and considering that the average human heart rate variability is in the range of 19-75 ms RMSSD, error rates close to or less than ~30 ms RMSSD can be considered acceptably accurate for distinguishing between broad HRV level groups. Our method shows good performance on datasets that are known to have low video compression and relatively less movement (PURE, UBFC-RPPG, VicarPPG, StableSet). Reasonable performance is also obtained on some datasets containing movement (MoLi-PPG and EatingSet); while good results are obtained on VicarPPG 2 in spite of subject movement. However, accuracy is very poor on the remaining datasets that either contain high compression noise (MAHNOB-HCI, COHFACE), or exhibit large/fast movements (VIPL-HR, ECG-Fitness).

5.3. In-depth Analysis

5.3.1. Effect of window length on heart rate computation

Considering the choice of window length over which heart rate is computed is important. In principle, estimating the average heart rate over larger time windows is an easier task than estimating instantaneous heart rates over shorter windows. This is because variations in the inter-beat intervals caused by falsely detected and missed peaks average out over a large window length, thereby giving the evaluation a higher tolerance to incorrect peak detections. To illustrate this, Figure 6 plots the factor by which the error rate changes when using different window lengths with respect to the long-term average heart rate error over the whole video. It can clearly be seen that error rates increase exponentially with decreasing window lengths for most datasets. The datasets less affected by this are the ones on which errors caused by other factors (like movement, compression, etc.) overshadow the effect of window length (e.g. ECG-Fitness, MAHNOB-HCI). This experiment illustrates the severity by which the chosen window size can affect the heart rate estimation accuracy and this should be taken into consideration during comparative evaluations.

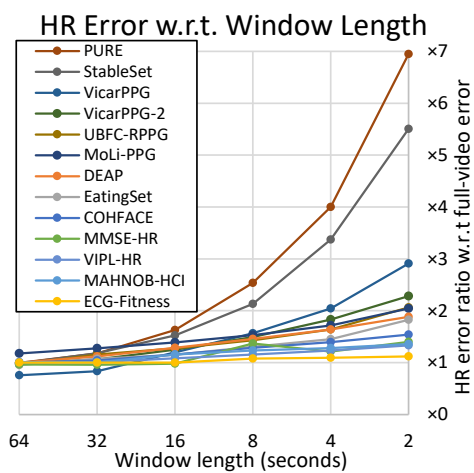

**Figure 6.** Heart rate estimation error rate (MAE) ratio with varying time-window lengths with respect to average long-term heart rate estimation error over the full video. The graph shows that the task of HR estimation becomes exponentially harder with decreasing window lengths. For example, error rate on PURE over 2-second window is 7 times worse than over the full video.

The rest of the experiments in this paper are performed with a window length of 16 seconds for heart rate estimation.

5.3.2. Effect of lighting conditions

An important factor that leads to attenuation of the underlying blood volume pulse in the extracted rPPG signal is the illumination/lighting conditions in the video. We test our method under various lighting conditions on datasets containing labelled illumination settings: COHFACE, VIPL-HR, ECG-Fitness, and MoLi-PPG; and the results are presented in Figure 7. We see that bright frontal lighting counter-intuitively leads to a degradation in performance in VIPL-HR. This is likely caused by pixel saturation, where the full colour depth of the camera cannot be exploited. Constantly flickering illumination from a computer monitor in MoLi-PPG also degrades the performance severely since such fluctuations can interfere with the pulsating component in the skin pixels. The colour temperature of the light also seems to have an influence as seen in ECG-Fitness, potentially due to dissimilar light wavelength absorption characteristics of the skin. In COHFACE, as can be expected, better performance is obtained when the room is evenly lit in comparison with uneven natural lighting. Finally, the method has good performance in the case of dark/dim lighting in MoLi-PPG and VIPL-HR.

5.3.3. Influence of subject movement

The physical movements of the subjects themselves can also introduce noise in rPPG extraction. This is essentially caused by the interaction of light on the observed skin pixels as it moves. The performance of our methods while subjects perform different kinds of movements are shown in Figure 8 on datasets containing labelled movement conditions: VIPL-HR, MoLi-PPG, ECG-Fitness, PURE, and VicarPPG 2. As per expectations, we observe that the rPPG method is most accurate when no movement is happening or when just the eyes are moving. All other kinds of movements degrade the performance somewhat, although the accuracy stay acceptable for some of them. The worst performance is observed when subjects perform large/sudden movements in multiple axis, as well as 'freestyle' head bobbing/dancing motion. In both MoLi-PPG and VicarPPG 2, an interesting observation is that vertical angular head movement (nodding) results in poorer accuracy while horizontal motion (head shaking) is handled quite well by the method. This could be due to the face modelling step being able to model side faces better than top/bottom looking faces.

5.3.3.1. Impact of rhythmic motion noise suppression. The impact of explicitly detecting and reducing the head movement noise in the signal via the rhythmic motion noise suppression component (Section 3.3) can be gauged with the help of an ablation study. The results of such a study on VicarPPG 2 and MoLi-PPG are shown in Figure 9. It can be seen that the addition of the motion noise suppression component in the rPPG pipeline reduces heart rate estimation errors significantly under most movement conditions. While the reduction in error is negligible when subject movement is low (conditions steady, stable, eye movement; ∼0.05 bpm), this gap in performance becomes large when subjects perform large intense head movements (∼1 to ∼3 bpm). The exception to this is the relatively small error difference for the left/right (MoLi-PPG) and tilting (VicarPPG 2) conditions, which can largely be attributed to poor AAM face fitting during movement resulting in incorrect head orientation measurement.

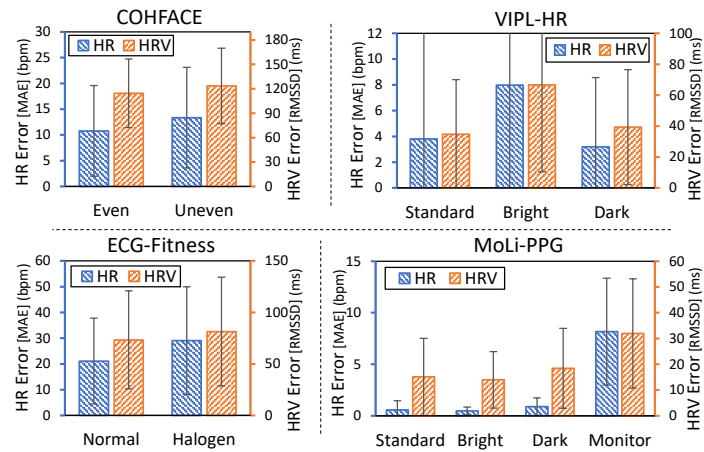

**Figure 7.** rPPG performance (MAE; error bars represent standard deviation) under different lighting conditions. Uneven illumination, oversaturation from by bright frontal lighting, yellowish glow of halogen lights, and the fluctuating monitor reflection, all somewhat degrade the performance of the rPPG method in comparison with standard ceiling lighting. However, the method performs well under dim/dark lighting.

A shortcoming of the rhythmic motion noise suppression method manifests itself when the primary frequency of head movement coincides with the heart rate of the subject. In such cases, while the suppression effect from low intensity head movements do not have a dominating effect against the heart rate frequency component, high intensity head movements can. They can significantly attenuate the heart rate component in the signal causing incorrect selection of the passband range of the narrow-bandpass filter, which can increase estimation errors. This phenomenon is observed in one of the movement conditions videos from VicarPPG 2 (subject #5).

5.3.4. Influence of facial activity

Another type of subject movement is the movement of facial muscles, commonly observed during talking, eating, or expressing emotions. Such movement leads to deformations in the shape of the face and stretching and moving of the skin, potentially leading to interference with the underlying BVP signal. We study the impact of facial movement on our rPPG method and present the results in Figure 10. The datasets included in this study are those containing labelled talking/eating condition (PURE, MoLi-PPG, VIPL-HR, EatingSet) as well as those exhibiting high facial expression activity (MMSE-HR). In all datasets, we see that talking leads to poorer performance as compared to when the face is static. Videos with higher facial arousal (an indication of facial expression activity) in MMSE-HR[‡] also result in a higher error rate for HR estimation. However, while eating, chewing motion does not seem to significantly influence accuracy. In fact, closer observation revealed that occlusion of the face caused when subjects take a bike/sip seem to be the larger cause of errors. The largest duration of facial occlusion happens when subjects take a sip (glass and hand covers the face), leading to a lower HR/HRV estimation accuracy.

5.3.5. rPPG in high HR ranges

To study how well the accuracy of the rPPG method spans over the range of heart rates, we can compare it's performance for subjects in a rested state versus when they are in a post-workout state. This can be seen in

---

‡　Facial expressions and arousal on MMSE-HR were obtained via an automated facial expression analysis tool called FaceReader [63].

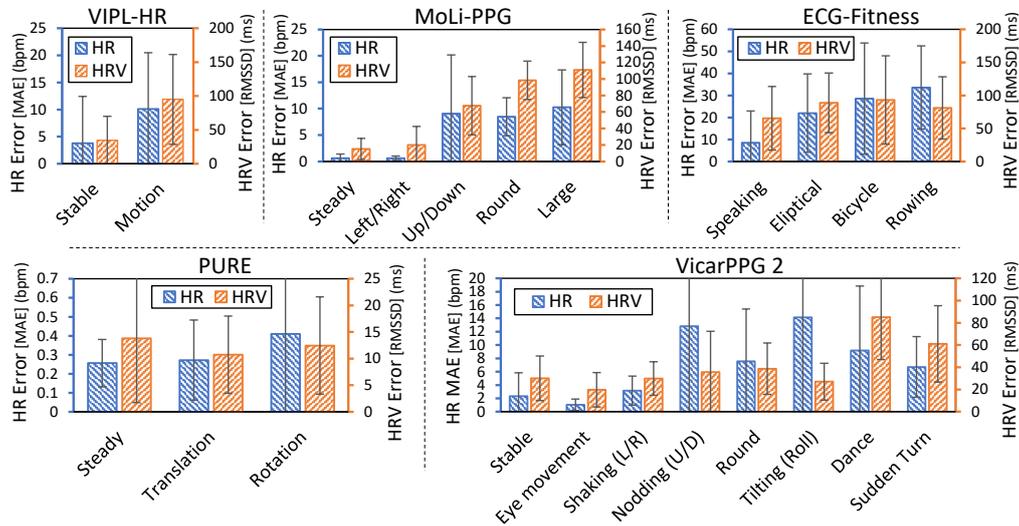

**Figure 8.** rPPG performance (MAE; error bars represent standard deviation) for different types of subject movements. Large/sudden multi-axis movement have the largest impact on accuracy. Interestingly, vertical head nodding motion seems to produce much higher error than horizontal head shaking motion of the same intensity.

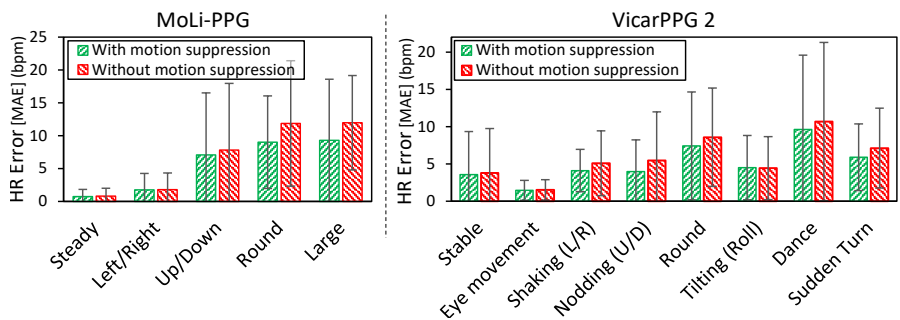

**Figure 9.** rPPG performance (MAE; error bars represent standard deviation) with and without the rhythmic motion noise suppression, for different types of subject movements on VicarPPG 2 and MoLi-PPG. In both datasets, motion noise suppression results in lower errors, especially when the subject movement is large.

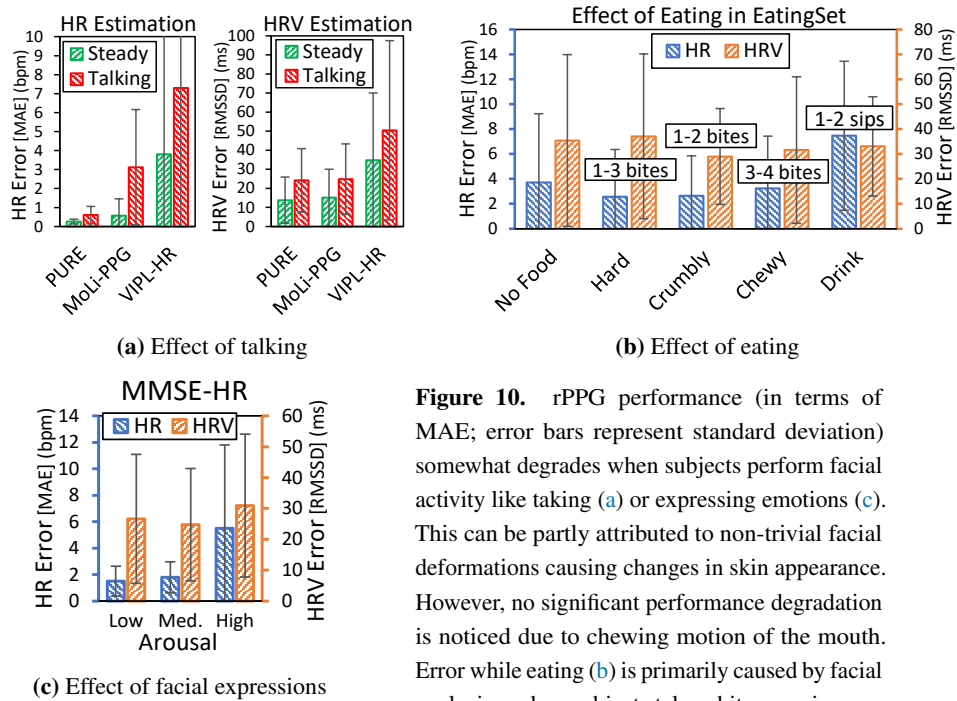

(a) Effect of talking

(b) Effect of eating

(c) Effect of facial expressions

**Figure 10.** rPPG performance (in terms of MAE; error bars represent standard deviation) somewhat degrades when subjects perform facial activity like taking (a) or expressing emotions (c). This can be partly attributed to non-trivial facial deformations causing changes in skin appearance. However, no significant performance degradation is noticed due to chewing motion of the mouth. Error while eating (b) is primarily caused by facial occlusion when subjects take a bite or a sip.

Figure 11b. Furthermore, we can explicitly group the videos from all datasets based on their average ground truth heart rate / HRV and measure the rPPG performance on them separately. Figure 11a plots the relative error rate ratio (MAE) per group for each dataset w.r.t. the average error over the whole dataset. It can be noticed in Figure 11b that while the error rates in the post-workout condition does seem to be marginally higher than the baseline condition, the proposed method performs with sufficiently good accuracy in all conditions. On VicarPPG, closer examination revealed that the variable frame-rate of some videos often drops sharply. This affects the estimation of higher HRs more severely as the Nyquist sampling frequency requirement is also higher. More generally, heart rate estimation in higher HR ranges appears to produce higher errors more often, as seen in Figure 11a. A contributing factor for this could be the presence of higher frequency noise in the same frequency range as the higher heart rates. For HRV, no such trends could be observed in relation to the HRV ranges.

5.3.6. Influence of video compression

Pixel-level noise caused by common video compression formats can be a major source of errors for rPPG methods. In principle, the cleanest signal is obtained from uncompressed video frames, but this can be impractical due to their large size. To study this further, we evaluate our methods on a range of video compressions levels and formats to determine the trade-off between rPPG accuracy and video size/bitrate. The results can be seen in Figure 12 for the PURE dataset. The result show that among lossy encodings (denoted by circle), H.265 format best retains the rPPG pulsating component in the skin pixels of the video, resulting in high HR estimation accuracy, while maintaining the lowest bitrate. For example, accuracy only drops by ∼0.3 bpm while the bitrate reduces by two orders of magnitude (26 Mbps to 0.26 Mbps). These results agree with the findings in [64]. Note that even under the lowest compression settings, these formats are not lossless and they result in a change in rPPG accuracy. Among the truly lossless codecs, FFV1 is able to encode the videos most efficiently. It results in zero drop in rPPG accuracy while reducing bitrates by almost an order of magnitude w.r.t raw videos. This can make rPPG dataset storage management much easier: for example, ECG-Fitness originally takes up 1.05 TB, but can be compressed with FFV1 to under 150 GB without losing any information.

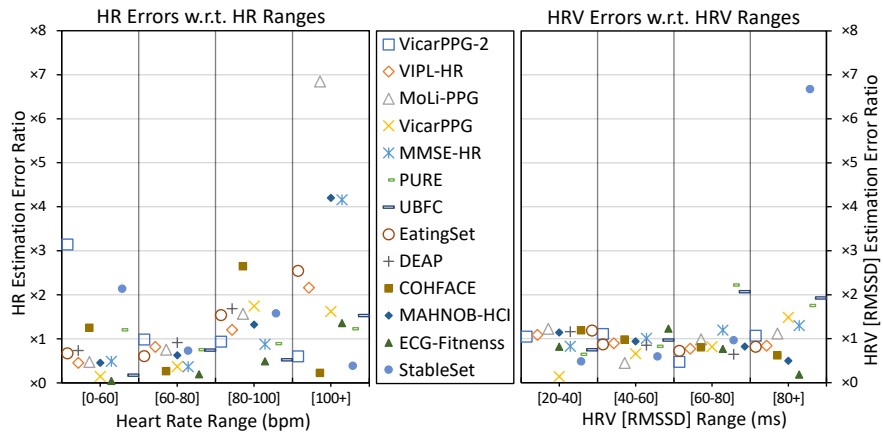

(a) rPPG performance in HR and HRV ranges

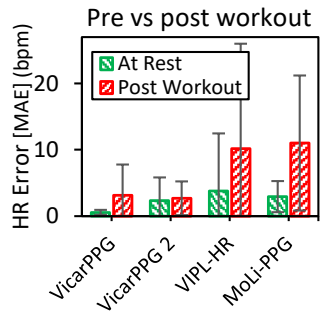

(b) Post-workout conditions

**Figure 11.** rPPG performance for (a) HR/HRV ranges in terms of MAE ratio w.r.t the average mean absolute error per dataset; and (b) rested vs post-workout conditions in terms of MAE (error bars represent standard deviation). With a few exceptions, HR estimation in higher ranges (including during post-workout conditions) is more often less accurate than in lower ranges.

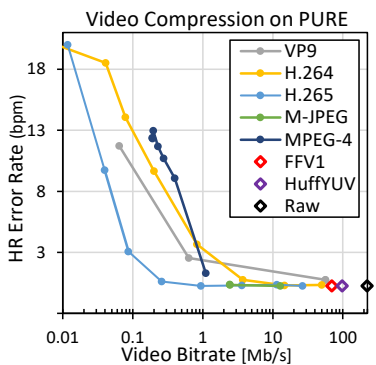

**Figure 12.** Effect of video compression on rPPG performance (MAE; PURE dataset). As videos are compressed to reduce their bitrate and storage size, rPPG accuracy decreases. H.265 offers the best trade-off between them among lossy encodings (denoted by circle). Among truly lossless encodings (denoted by diamond), FFV1 stores videos most efficently without affecting rPPG accuracy at all.

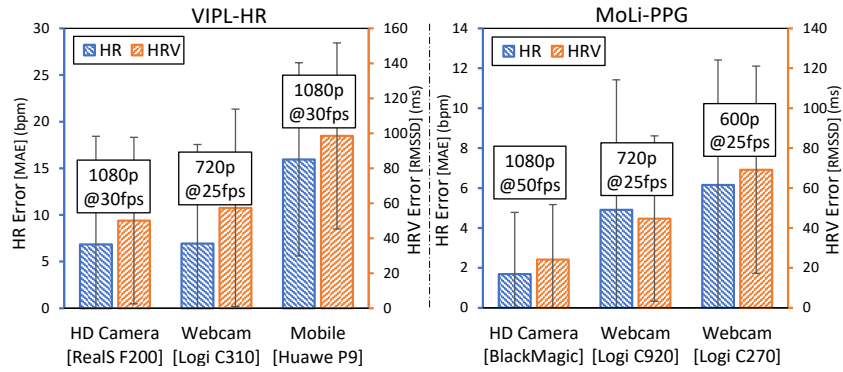

**Figure 13.** rPPG performance (MAE; error bars represent standard deviation) w.r.t camera type. HD cameras produce the best results likely due to their superior sensor. Modern webcams perform fairly well too.

| Face Finding & Modelling (ms) | Skin Pixel Selection (ms) | rPPG Algorithm (ms) | Total (ms) | Frame Rate |
|---|---|---|---|---|
| 31.89 ± 17.2 | 0.43 ± 0.2 | 0.56 ± 0.2 | 32.88 ± 18.8 | ~30.4 fps |

**Table 4.** The processing speed of individual components of the proposed method's pipeline and the total frame rate (640×480 pixel input) on an Intel Xeon E5 CPU. The main bottleneck is face finding and modelling, while the rest require negligible time.

#### 5.3.7. Effect of camera type

A closely related factor that also affects rPPG analysis is the camera type, which determines the signal acquisition quality. Figure 13 provides a comparison of HR and HRV accuracy for different camera types used in VIPL-HR and MoLi-PPG datasets. As can be seen, the HD cameras provide the best performance in both datasets, likely due to their superior sensor, low internal compression, and higher frame-rate. However, we can also see that a modern webcam can closely match the performance of an HD camera under realistic condition (especially in VIPL-HR). The mobile front camera (Huawei P9) produces the worst performance in VIPL-HR, likely caused by it's significantly lower video acquisition quality.

#### 5.3.8. Processing Speeds

For real time application, processing speed is just as vital as prediction accuracy. The average CPU processing times of our method and its individual components are listed in Table 4 (on an Intel Xeon E5 processor). The method performs the full analysis with a good real-time speed for a video resolution of 640×480 pixels. The only bottleneck is the face finding and modelling step, which is modular w.r.t the rPPG pipeline and can be swapped out for a faster implementation.

### 6. Discussion and Conclusions

We were able to obtain successful and promising results from our appearance modelling and signal-processing based rPPG method. The results show that this unsupervised method achieves high accuracies, matching or surpassing state-of-the-art on six public datasets: PURE, VIPL-HR, MoLi-PPG, VicarPPG, UBFC-RPPG, and MMSE-HR. In fact, the accuracy of our method for heart rate analysis is in the same range or beyond several fully supervised deep learning methods, albeit without any rPPG specific training or fine-tuning. We also surpass all existing methods for heart rate variability estimation and set some of the first benchmarks for heart rate variability analysis on these datasets.

Through an exhaustive 13-dataset evaluation (including release of a new public dataset: VicarPPG 2), the strengths and weaknesses of our method were highlighted. We showed that the proposed method handles most realistic variations in illumination, movement, and facial activity well. This can be attributed to the appearance modelling and noise suppression steps in the pipeline. However, certain combinations and extreme cases of these conditions proved challenging: overtly bright or flickering lighting and large head and body movements (e.g. during exercising). Our study provided some unique insights about the rPPG analysis in terms of performance while eating and emoting facial expressions: chewing motion during eating did not result in larger errors, while HR analysis during high facial arousal proved marginally challenging.

We also explicitly studied the impact of additional recording factors like video compression and camera type: H.265 and FFV1 emerged as clear winners in terms of preserving plethysmographic information in the skin pixels efficiently; higher quality rPPG signal can be obtained from HD cameras, but modern webcams also provide good results. High video compression noise was observed to be a clear limitation of our signal-processing method, especially in comparison with deep learning based method. Several deep learning methods have shown good results on such datasets, while they fail to match our method in cases with lower compression. This could be because the deep network is able to learn the spatial patterns of this compression noise and filter them out. In contrast, in lower compression cases, our prior domain knowledge assumptions perform more accurately. While this makes our method well suited for modern videos, deep learning might be better suited for processing archival videos, often subject to higher compression.

Finally, we demonstrated how the ground truth PPG and ECG signals provided with most datasets can be highly noisy, leading to incorrect peak detection and resulting in substantial miscalculation of heart rate and heart rate variability measures. To tackle this, we introduced the CleanerPPG set: a collection of hand-cleaned ground truth peaks for 13 major public datasets. Using this ground truth ensures a fairer and more accurate evaluation. This set is intended to continuously grow with community support.

**Author Contributions:** Conceptualization, A.G. and M.B.; methodology, A.G. and M.B.; software, M.B.; validation, A.G. and M.B.; formal analysis, A.G. and M.B.; investigation, A.G. and M.B.; resources, J.vG.; data curation, A.G. and M.B.; writing–original draft preparation, A.G.; writing–review and editing, A.G., M.B., J.vG.; visualization, A.G.; supervision, J.vG.; project administration, A.G.; funding acquisition, M.B. All authors have read and agreed to the published version of the manuscript.

**Funding:** This work is partly supported by the Dutch Research Council (NWO) under the Citius Altius Sanius perspective programme P16-28 Project 4.

**Acknowledgments:** The authors would like to thank Roelof Lochmans and dr. Daniël Lakens from the Human-Technology interaction at Eindhoven University of Technology for their valuable contributions towards recording of the StableSet dataset.

**Conflicts of Interest:** The authors declare no conflict of interest.

## Abbreviations

The following abbreviations are used in this manuscript:

| | |
|---|---|
| rPPG | Remote Photo-plethysmography |
| PPG | Photoplethysmogram/graphy |
| BVP | Blood Volume Pulse |
| ECG | Electrocardiogram/graphy |
| HR | Heart Rate |
| HRV | Heart Rate Variability |
| IBI | Inter-Beat Interval |
| MAE | Mean Absolute Error |
| bpm | beats per minutes |
| RMSSD | Root Mean Square of Successive Difference |
| LF/HF | Low Frequency / High Frequency |
| FFT | Fast Fourier Transform |
| fps | frames per second |